# Integrating Machine Learning with Discrete Event Simulation for Improving Health Referral Processing in a Care Management Setting

Mohammed Mahyoub[1, 2]

[1]Systems Science and Industrial Engineering Department, State University of New York at Binghamton, Binghamton, NY 13902, USA

[2]Virtua Health, Marlton, NJ 08053, USA

mmahyou1@binghamton.edu

# Integrating Machine Learning with Discrete Event Simulation for Improving Health Referral Processing in a Care Management Setting

**ABSTRACT**

Post-discharge care management coordinates patients' referrals to improve their health after being discharged from hospitals, especially elderly and chronically ill patients. In a care management setting, health referrals are processed by a specialized unit in the managed care organization (MCO), which interacts with many other entities including inpatient hospitals, insurance companies, and post-discharge care providers. In this paper, a machine-learning-guided discrete event simulation framework to improve health referrals processing is proposed. Random-forest-based prediction models are developed to predict the LOS and referral type. Two simulation models are constructed to represent the as-is configuration of the referral processing system and the intelligent system after incorporating the prediction functionality, respectively. The LOS prediction models for the three affiliated hospitals performed well ($H_1$: MAE = 1.9 ± 0.12, $R^2$ = 0.67 ± 0.04; $H_2$: MAE = 2.35 ± 0.07, $R^2$ = 0.64 ± 0.02; $H_3$: MAE = 2.52 ± 0.0.36, $R^2$ = 0.50 ± 0.09). The referral type prediction model performed well across all hospitals, too ($H_1$: Accuracy= 90.15 ± 0.83; $H_2$: Accuracy= 88.59 ± 1.66; $H_3$: Accuracy= 89.35 ± 1.84). By incorporating a prediction module for the referral processing system to plan and prioritize referrals, the overall performance was enhanced in terms of reducing the average referral creation delay time (RCDT) by about 45% compared to the baseline model. This research will emphasize the role of post-discharge care management in improving health quality and reducing associated costs. Also, the paper demonstrates how to use integrated systems engineering methods for process improvement of complex healthcare systems.

## 1. INTRODUCTION

The U.S. healthcare system has been experiencing many issues including patient rehospitalization, which accounts for very high costs (Jencks et al., 2009). For example, $15 billion is spent on Medicare readmissions (Hackbarth, 2009). Readmissions can be prevented if suitable interventions are applied. Such interventions can be manifested as transferring patients from hospitals to appropriate post-discharge facilities or services including skilled nursing facilities (SNF) and health home services (HHS). The process of transferring patients from hospitals to post-discharge destinations is usually referred to as health referrals, especially in the post-discharge care management setting adopted in this paper. An important criterion of a smooth, successful patient transition is the availability of the health referral upon discharge from the hospital. Therefore, process improvement initiatives should be implemented to ensure the availability of referrals upon the patient's discharge. To carry out a successful process improvement project, first, the system under investigation should be understood. The following discussion will shed light on the complexity of the healthcare system including post-discharge care management and the role of systems thinking in improving its processes. Also, the research problem will be explicitly defined.

The healthcare system is concerned with maintaining the health of the population in the most optimized way. Healthcare process improvement initiatives work on improving the system by pursuing the three dimensions of the "Triple Aim" (Berwick et al., 2008). The "Triple Aim" ensembles three goals: (1) improving the patient experience of care, (2) enhancing population health, and (3) reducing healthcare costs per capita. The system's complexity can make such optimization efforts very difficult; the healthcare system in the U.S. is very complex (Tien and Goldschmidt-Clermont, 2009). The complexity arises from the system's structure and function. It consists of many components including patients, providers, government bodies, insurance companies, and care management organizations. Additionally, these components are interrelated, having intricate relationships, and cross-functionality. The healthcare system has four main functions: finance, insurance, care delivery, and payment (Shi and Singh, 2014). Looking further into each component, there are many subcomponents, and they are also interrelated and have cross-functionality. So, it is obvious now that a comprehensive understanding of the overall system is urgently needed to enhance its performance and being able to deliver high-quality, low-cost services.

A noticeable reflection of the overall healthcare system's complexity is manifested in care management. Care management is concerned with providing care services to specific groups of patients, who are enrolled in particular plans or programs based on their health conditions and negotiating prices or payment arrangements with providers (Shi and Singh, 2014). The care management field has played an important role in transforming healthcare services delivery (Shi and Singh, 2014). To illustrate, a Managed Care Organization (MCO) is concerned with transforming the current health system whose entities tend to work in silos into an integrated platform, which incorporates finance, insurance, care delivery, and payment into one organization. The MCO works as a coordinator responsible for securing needed health services of enrollees and negotiating with insurance companies and providers to pay for those services at discounted fees achieved by mutual contracts between the care management organizations, providers, or insurance companies. Care management is considered a national priority to improve the quality of care by achieving safety, effectiveness, efficiency, patient-centeredness, timeliness, and equity (Carroll, 2002). The MCO establishes many specialized programs and plans for chronic diseases and severe health conditions (e.g., diabetes and heart disease). Often, a specialized care management program is devoted to managing patients after being discharged from hospitals; the program is usually referred to as post-discharge care management.

Post-discharge care management is a very important component within the model of transitional care. This can be attributed to the fact that discharged patients are associated with high risk because of the strong correlation between adverse events and patient discharge (Moore et al., 2003; Forster et al., 2003). Transitional care is concentrated upon preparing patients for a smooth transition from one care setting (e.g., inpatient hospital) to another setting (e.g., post-discharge facility). To ensure a smooth transition, transitional care specialists perform discharge planning, discharge preparation, care coordination, and case management (Weis et al., 2015). The delivery of post-discharge services is a very complex process, which involves the coordination of many entities including patients, hospitals, insurance companies, and post-discharge care facilities (e.g., HHS and SNF). This coordination is very crucial to reduce readmissions and unnecessary emergency department visits (Bisognano and Boutwell, 2009; Kilcup et al., 2013). Besides, this is very crucial in improving the patient's health profile and preventing catastrophic conditions (Allen et al., 2002). Post-discharged care management (i.e., health referral processing) is performed by a specialized unit within the MCO (see Figure 1). Health referrals should be made available upon patients' discharge. However, health referral processing can be hindered by many factors including, the uncertainty of patient post-discharge destinations (i.e., referral type), discharge dates, and inefficient planning at the MCO side.

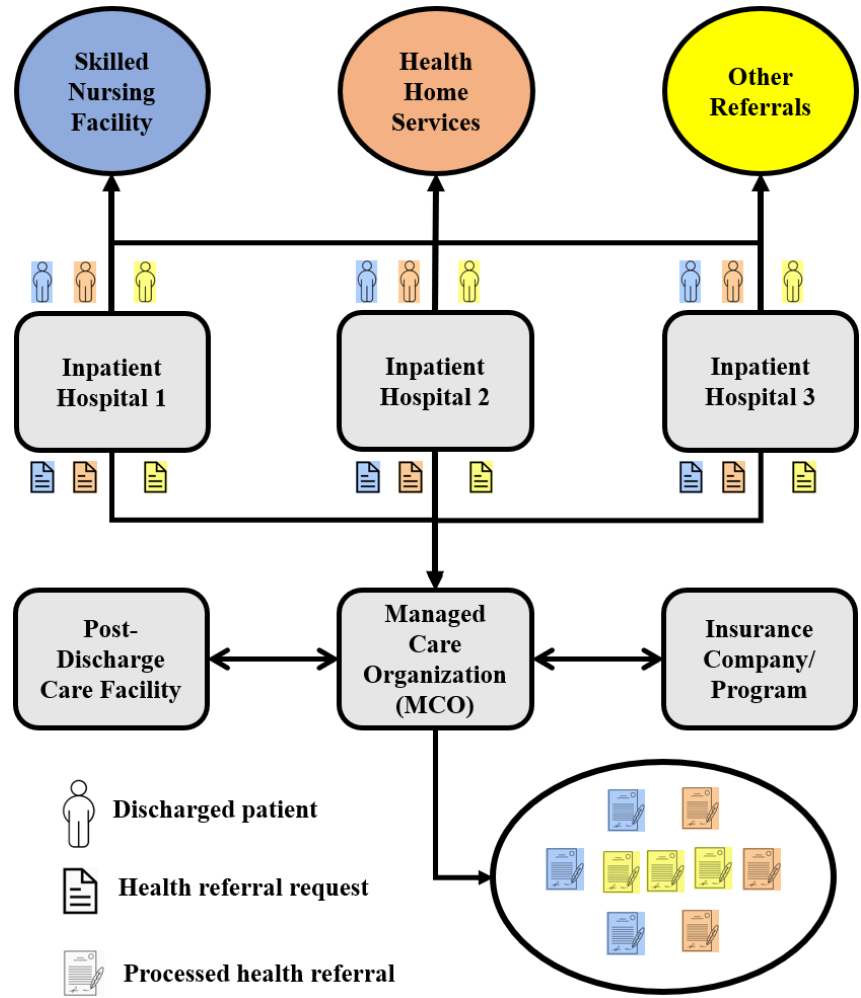

Figure 1. Health referral processing coordinated system flowchart

The objective of this paper is to develop and implement a machine-learning-guided discrete event simulation framework for improving health referral processing. The main goal is to reduce the health referral delays exerted by the uncertainty of referral type, LOS, and the inefficient planning at the MCO side (i.e., lack of prioritization criteria). The framework is composed of two parts: prediction and simulation. The prediction part aims at estimating the referral volume and the simulation part represents the coordinated system of health referral processing. The proposed process improvement framework is centered around incorporating a prediction module to predict patient discharge date and referral type to allow for early demand estimation, which is introduced to the simulation model. The latter can be used for early planning, processing, and prioritization. Integrating machine learning with discrete event simulation provides the advantage of testing the proposed approach. The discrete event simulation part models the referral request journey from the inpatient hospital to the MCO. This paper is based on thesis work submitted to obtain a master's degree (Mahyoub, 2020).

The paper is structured as follows: Section 2 reviews the literature related to the scope of this paper. Section 3 introduces the methods and materials used to conduct this study. Section 4 presents the results of the prediction models, baseline discrete event simulation model, and the machine-learning-guided simulation model. Section 5 discusses the results and their practical implications. Finally, Section 6 summarizes the extracted insights, limitations, and opportunities for future work.

## 2. Literature Review

This section will present an overview of related studies discussing machine learning or data mining techniques for predicting LOS and post-discharge destination, simulation modeling in healthcare, and integrating machine learning with simulation modeling.

Data mining is a knowledge discovery approach and has been used extensively in many settings to take advantage of available data, extract useful insights, and build predictive models to enhance business processes. Usually, data mining techniques are categorized into two types: descriptive and predictive (Yoo et al., 2012). Predictive data mining was used in this paper. The key difference between the two types is that predictive analytics is applied to problems in which a response variable is clearly defined. The response variable might be continuous or categorical. Based on the latter categorization of the response variable, data mining techniques can be further classified into regression (i.e., for problems having a continuous response variable) and classification (i.e., for problems having a categorical response variable) (Bellazzi and Zupan, 2008). In the healthcare sector, data mining is becoming increasingly essential to benefit all parties involved in the healthcare industry (Koh and Tan, 2011).

In care-management-related settings, predictive data mining has been adopted for specific purposes such as population stratification and identification of eligible patients for care management. This has been achieved through the prediction of patient health status (Dominic et al., 2015; Shouval et al., 2014; Sun et al., 2014) and medical costs (Sushmita et al., 2015; Shenas et al., 2014; Bertsimas et al., 2008). However, there is a lack of applying data mining techniques for improving the operational performance of the managed care organizations (MCO). Therefore, there is a vital need for studies showing successful applications of machine-learning-based predictive modeling in enhancing the performance of MCO operations including post-discharge management.

In this paper, machine-learning-based predictive models, i.e. predictive data mining algorithms, were used in the first part of the health referral processing improvement framework to estimate the demand of the MCO based on discharge data at the affiliated inpatient hospitals. The demand quantity (i.e., the number of patients discharged on a given data) was estimated by predicting the length of hospital stay (LOS) of admitted patients using a random-forest-based regression model. Predicting the LOS has been the focus of many studies in the literature because of its importance in hospital resource planning and hospital performance evaluation (Koh and Tan, 2011; Zolbanin et al., 2020; Lu et al., 2015). Various machine learning algorithms were utilized to predict LOS of different groups of patients such as support vector machine, decision tree, neural network, random forest, and other methods (Al Taleb et al., 2017; Veloso et al., 2014; Hachesu et al., 2013). Nevertheless, most of these studies focused on predicting the LOS for specific groups of patients (e.g., chronically diseased patients). This is not useful for post-discharge planning, which needs a comprehensive predictive modeling approach in which all admission data is considered to make the planning more plausible.

The problem of predicting post-discharge destinations, i.e. referral type, for specific groups of patients was addressed in some research studies. Pereira et al. (2014) developed a discharge planning prediction model based on a logistic regression approach for patients with severe stroke. A binary classification problem was solved to predict the post-discharge destination (i.e., home or not home) of rehabilitation patients using signal detection analysis (Miyamoto et al., 2008). Another study predicted the discharge destination for patients after undertaking laryngectomy (Panwar et al., 2018). Finally, Sivertson et al. (2010) examined the validity of predicting the post-discharge destination for femoral neck fracture patients. The discussion above emphasized the fact that like predicting LOS, most of all the reviewed studies focus on a particular group of patients. Another limitation is the lack of multi-class classification models for predicting more than two classes (i.e., post-discharge destinations). Most of the studies above estimated whether the patient will be transitioned to a specific post-discharge destination or not. This does not reflect the diverse nature of destinations.

The second part of the referral processing improvement framework, propped in this paper, is simulation modeling. A Discrete Event Simulation (DES) approach was adopted. DES is a computerized method of representing a real-world system (e.g., healthcare system) and its associated operations to imitate the changes over time to provide objective insights based on evidence to aid decision-makers in developing operational solutions (Hamrock et al., 2013). In the past years, DES has been used to address various issues in healthcare settings including improving patient flow (Parks et al., 2011; Marshall et al., 2005), bed management (Landa et al., 2014; Cardoen and Demeulemeester, 2007), and performance modeling in general (Günal and Pidd, 2010). However, there is a notable shortage of studies on applying DES for process improvement in care management settings (e.g., health referral processing).

The integration of machine learning and simulation for process improvement is a new immerging area of research that has the potential to model complex systems and enhance the performance of its processes. Elbattah and Molloy (2016) developed a machine-learning-aided simulation model for discharge planning. In their model, from a discrete event simulation perspective, the patient's journey through the hip fracture care system was modeled. Then, the developed predictive models were used to make predictions on the inpatient length of stay and discharge destination of the simulation-generated patients. Their overall model was used to predict the demand for a certain geographical area. The focus of the research problem and the approach adopted in this paper is different from those implemented by Elbattah and Molloy (2016). In this paper, integrating machine learning with simulation was used to model and enhance the performance of a health referral processing system. The machine learning part was used to predict the demand of the MCO. Then, this information was introduced as an input to the DES model to improve the performance of the overall system under the predicted demand initiated by affiliated inpatient hospitals.

Including the study discussed above, several other studies emerged in the literature that outline and adopt the idea of coupling artificial intelligence with simulation to model complex systems. A recent literature review study examined the use of DES in conjunction with big data analytics (Greasley et al., 2019). For instance, Aqlan et al. (2017) integrated machine learning (i.e., artificial neural network) with simulation for defect management in manufacturing environments. Another study presented an outline for the cooperation of machine learning and simulation for performance analysis of complex systems (Kim and Kim, 2019). De La Fuente et al. (2018) demonstrated the integration of DES and deep learning in a bank credit approval process setting. The major motivation for integrating machine learning with simulation is to increase simulation models' validity by considering real data modeling and take advantage of available the big scale data. Yet, the application of this interesting methodology is in its infancy and many studies are needed to uncover its true potential, especially in healthcare settings.

In general, there is a shortage of studies concerned with using integrated approaches or frameworks to model such complicated relationships among the healthcare system entities. In consequence, there is an inevitable need to develop integrated frameworks to model the relationship between several components of the healthcare system and their interrelated functions. This type of approach will contribute significantly to solidify the understanding of the overall healthcare system, thus identifying opportunities for improvement and efficient decision making. Specifically, there is an extreme lack of research on using systems engineering tools to improve the care management process. This paper introduces an integrated and data-driven systematic approach to improve health referral processing in a care management setting.

## 3. Methodology
### 3.1 Research Framework

In this paper, a machine-learning-guided simulation framework for improving health referral processing is developed and applied to a case study of a leading Managed Care Organization (MCO). The framework is composed of two core parts: prediction and simulation (see Figure 2). Patient discharge data is fed into the prediction models, which are developed using a random forest algorithm. A regression and multi-class classification problem are solved to predict the patient's LOS and referral type, respectively. LOS prediction enables the estimation of the patient's discharge date, thus the demand of the MCO is estimated as well (i.e., the number of referral requests on a particular day). Then, the predicted demand and health referral processing data (e.g., process workflow, staff level, and referral processing time) are introduced to a discrete event simulation model to evaluate the baseline model performance and future status performance after implementing the proposed approach. The proposed approach uses the predicted data to prioritize requests based on the discharge date and allow for early referral requests by hospitals once the referral type is predicted.

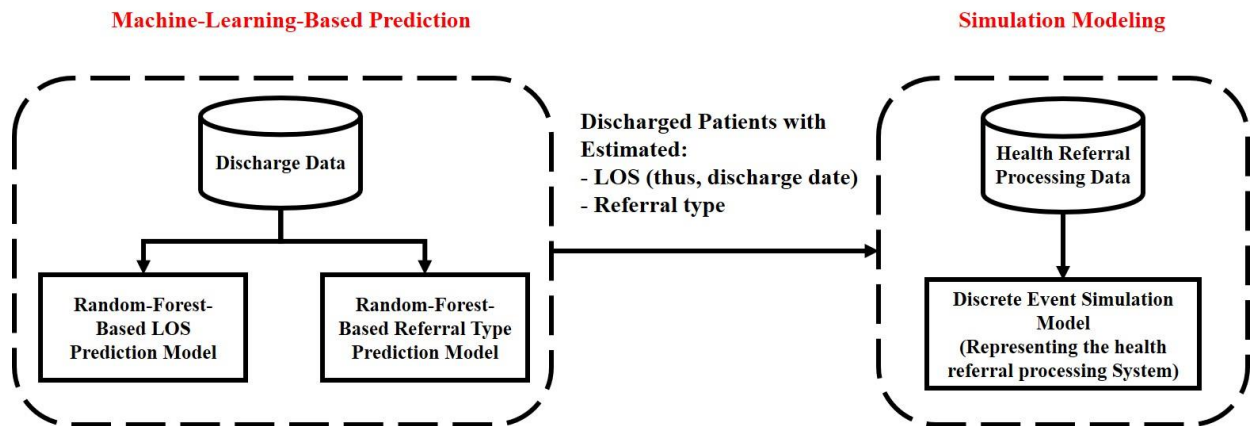

Figure 2. Machine-Learning-Guided Simulation for Improving Health Referral Processing Framework.

**3.2 Predictive Modeling**

Figure 3 depicts the predictive modeling research steps. As the first step in building the prediction models, discharge data were collected from the New York Department of Health (NYDOH) website (NYDOH, 2019). This research is concerned with three inpatient hospitals, which are affiliated with the targeted MCO. So, the data for each of these hospitals were extracted. Data cleaning was applied to prepare the data for subsequent research phases. Categorically encoded data was reencoded using numerical values to be feasible for the random forest algorithm. Outliers were removed from continuous variables (e.g., LOS). After that, the data was normalized. At this point, the data is preprocessed and ready for predictive modeling. Data was prepared for both problems: the regression and multi-class classification. To illustrate, the response variable for each problem is defined separately and the associated prediction model was coded separately as well. In terms of technical predictive modeling, first, a baseline model is built, hyperparameters are tuned, and finally, the enhanced model is evaluated. An additional step of balancing the data is adopted in the classification modeling branch. The last step in the predictive modeling part was to estimate the discharge volume (i.e, MCO demand) for a month, which was introduced to the simulation model.

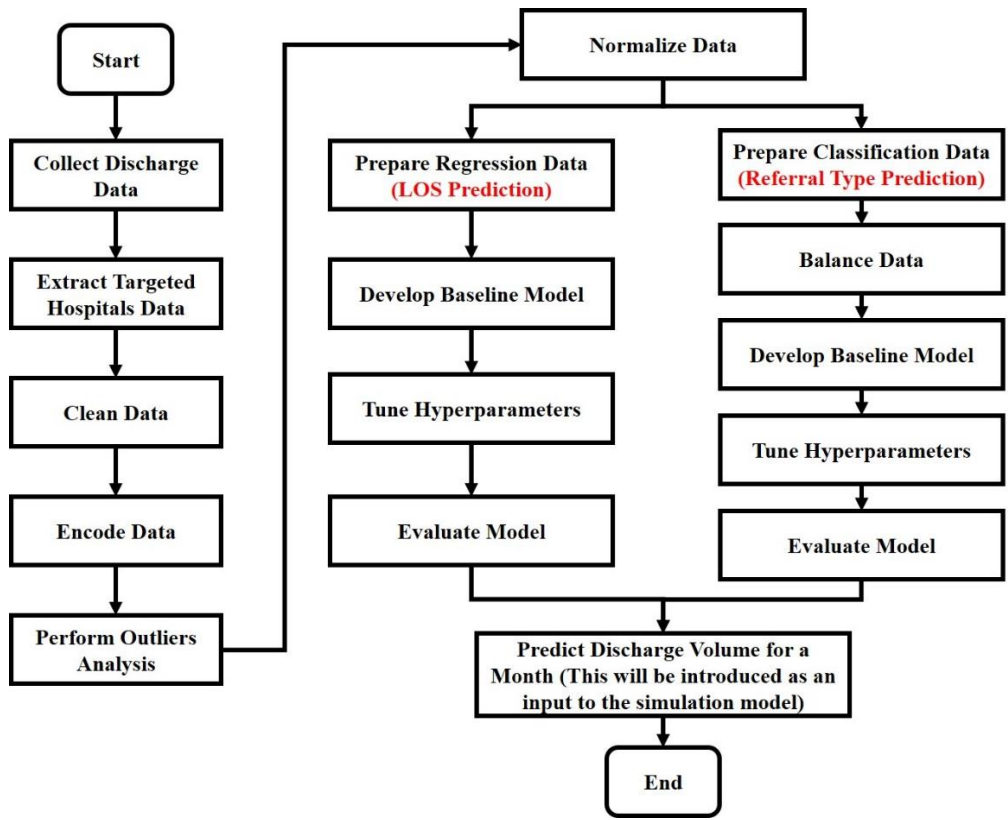

Figure 3. Predictive modeling flowchart

### 3.2.1 Data Description

The discharge data were collected for targeted inpatient hospitals from the NYDOH website (NYDOH, 2019). The patient discharge data contains details on patient characteristics, treatments, diagnoses, health risk measures, and charges information. The focus of this research is on a leading MCO in New York, so only the data of the MCO affiliated hospitals were extracted. Also, all irrelevant variables were excluded. This includes the Health Service Area, Hospital County, Facility ID, etc. The remaining viable variables were used as features and response variables for the prediction models. The features are Age Group, Gender, Race, Ethnicity, Type of Admission, AHRQ Clinical Classification Software (CCS) Diagnosis Category Code, AHRQ Clinical Classification Software (CCS) ICD-9 Procedure Category Code, All Patient Refined Diagnosis Related Groups (APR-DRG) Code, All Patient Refined Major Diagnostic Category (APR MDC) Code, APR Severity of Illness Code, APR Risk of Mortality, and Source of Payment. A detailed description of each of these variables is available on the original website (NYDOH, 2019) or in the thesis work associated with this article (Mahyoub, 2020). This paper aims to predict the demand for the MCO by predicting the patient's LOS and referral type. Therefore, two prediction problems were solved separately. The response variable of the regression model is patient LOS. The response variable of the multi-class classification model is the patient referral type (i.e., post-discharge destination). Three referral types were considered: SNF, HHS, and Other. It should be mentioned that LOS was also used as an additional feature to the referral type prediction model.

### 3.2.2 Data Preprocessing

Data preprocessing entails several steps. First, data cleaning by removing any invalid data instances is implemented. There were negligible corrupted data rows compared to the overall size of the dataset. Therefore, corrupted data instances were eliminated. Second, most of the original data variables are categorical and are filled with text labels corresponding to each level. For instance, for Gender, the levels are "Female" and "Male". Therefore, the data was encoded into numerical values for later incorporation into the machine learning algorithms. Usually, most of the mathematical and logical operations within a particular machine learning algorithm requires numerical data. The encoding process of text labels to corresponding numerical values was achieved using the conventional method of associating a number with a specific class. For instance, the multi-class classification model target variable, referral type, was transformed according to the following logic. SNF was replaced with 0, HHS was replaced with 1, and other referrals class was replaced with 2. Another important data processing step for machine learning algorithms input data is scaling feature variables (and sometimes target variable) to be in the same numerical range (e.g., 0 to 1). This will ensure all data values have the same contribution to the prediction model during training. To illustrate, feature variables with large numbers compared to other features might affect the model ability to generalize as it becomes biased or skewed by the large values. Thus, data were normalized using the Min-Max approach (Al Shalabi, 2006). After conducting an exploratory data analysis, it was shown that the original dataset is highly imbalanced considering the three referral types of SNF, HHS, and Other. To remedy this issue, the Synthetic Minority Oversampling Technique (SMOTE) algorithm was used to balance the dataset. SMOTE was chosen because it can create a generalized decision region for the minority class (Chawla et al., 2002). An equivalent problem for imbalanced data in regression modeling (i.e., when the target variable is continuous) is the presence of outliers. Outliers can skew model parameters and reduce its ability to accurately predict the target variable. Consequently, the ROUT method was used to remove outliers from the LOS variable. ROUT is a method that combines robust regression and outlier removal (Motulsky and Brown, 2006).

### 3.2.3 Prediction Models Development

The main objective of this paper is to propose and implement a machine-learning-guided simulation model to improve health referral processing in a care management setting. Speaking about the machine learning part, two prediction models were developed: a regression model for predicting patient LOS and a multi-class classification model for predicting the patient referral type. In this paper, a random forest was considered for both prediction problems. Several machine learning algorithms were examined including neural network and k-nearest neighbors. However, the random forest proved to outperform all other algorithms (Mahyoub, 2020). The focus of this paper is on the overall process improvement framework. So, only the random forest algorithm will be included in the following discussion. A brief description of the algorithm is presented below.

A random forest (RF) classifier is an ensemble learning method that fits several decision tree classifiers on randomly selected subsets of the targeted dataset (Breiman, 2001). In general, this ensemble method tends to combine and average the predictions of several base decision trees so that generalizability and robustness are improved compared to a single decision tree. Combining several base learners to improve the overall ensemble model performance in prediction is called boosting (Freund and Schapire, 1997). RFs follow the same principle as the boosting method of combining base learners (i.e., decision trees). However, boosting works by fitting week estimators (e.g., shallow trees). On the other hand, RF usually sums up the prediction power of fully grown trees. The random forest algorithm can also be used for solving regression problems in which the response variable is a continuous variable such as patient LOS in this study (Breiman, 2001; Cutler et al., 2012). RF can be also used for solving regression problems (Breiman, 2001; Cutler et al., 2012). Prediction models were built in Python using Scikit Learn (Pedregosa et al., 2011). Hyperparameters were tuned using the sequential model-based optimization implemented in the Hyperopt library in Python (Bergstra et al., 2015). It should be mentioned that each affiliated hospital's data was modeled separately to capture the independent referral requests in the real system.

To evaluate the performance of the random forest models, experimental results were collected using the 10-fold cross-validation approach (Kohavi, 1995). Several evaluation metrics were selected. Accuracy, Sensitivity, Specificity, and Area under the Receiver Operating Characteristic Curve (AUROC) were selected for evaluating the multi-class classification model. The reader can be referred to Tharwat (2020) for a detailed description of these metrics. The regression model evaluation was accomplished by determining the following metrics: Mean Square Error (MSE), Mean Absolute Error (MAE), and Coefficient of Determination ($R^2$).

## 3.3 Simulation Modeling

To study the current configuration of the system and test the effectiveness of the proposed approach in terms of referral processing and delay times, a discrete event simulation model (DES) was developed as a representation of the integrated system of the MCO and its affiliated hospitals. Simulation modeling was accomplished by following several steps (see Figure 4). The health referral processing related data was collected from the repository of a leading MCO. The collected data was cleaned and prepared for further research steps. Outliers analysis was performed. Then, data were fitted to the most appropriate probability distribution. After that, a baseline model was developed to evaluate the performance of the current configuration of the system. The baseline model was verified and validated against historical data to make sure the model is representative of the system in terms of the research goals. Following that, the performance of the baseline model was evaluated to be compared with the proposed model (i.e., machine-learning-guided DES). The proposed model was developed by altering the logic and some parameters of the baseline model. The MCO demand estimated by the prediction models was collected to be introduced to the ML-Guided DES model. Finally, the proposed model was evaluated and compared to the baseline model to measure the effectiveness of incorporating the prediction models.

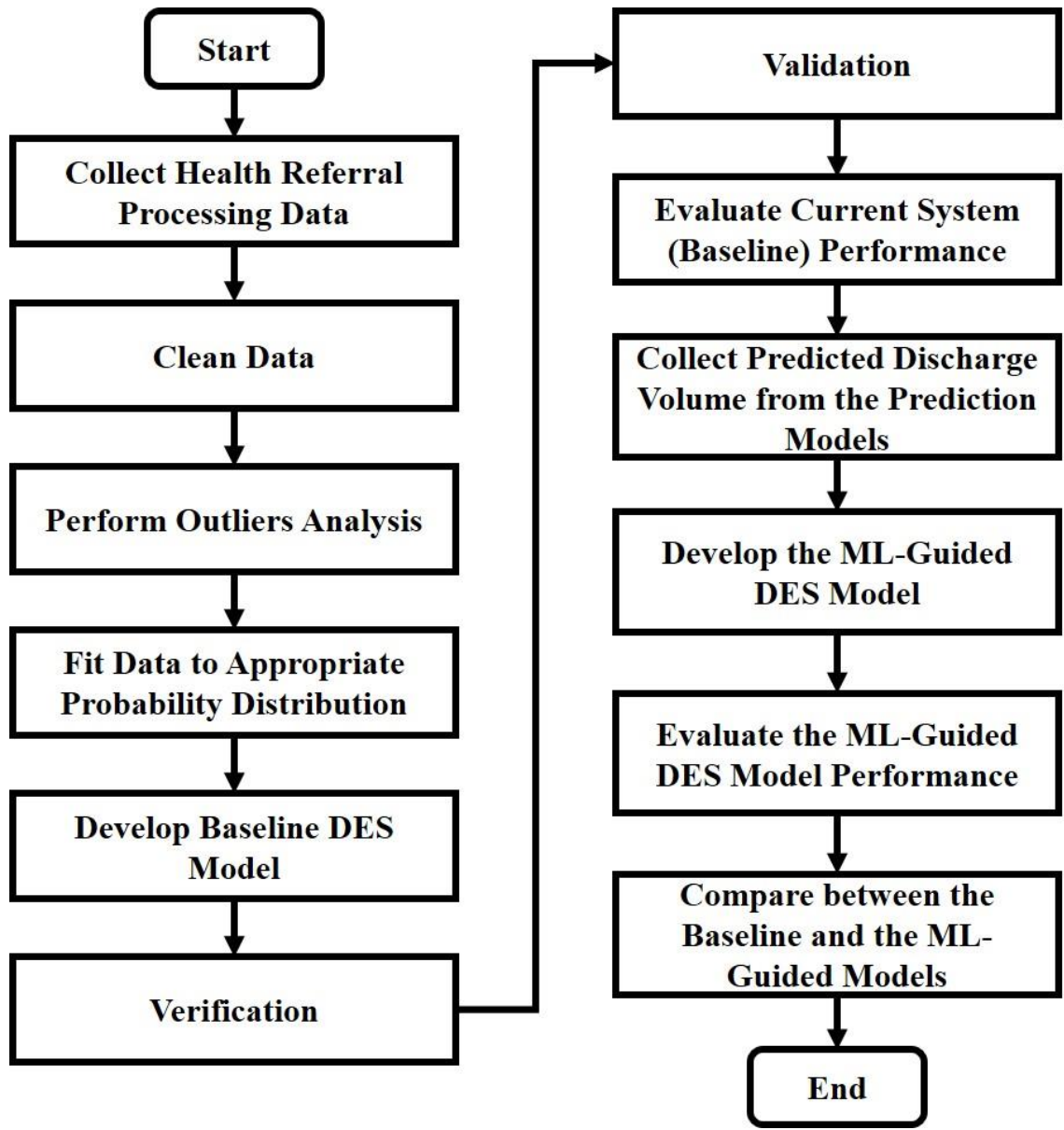

Figure 4. Simulation Modeling Flowchart

### 3.3.1 Data Description and Preprocessing

To increase the validity of the DES model, which is a representation of the health referral processing system, relevant data related to referral processing such as referral arrival rate per day, referral request processing time (at the hospital side), MCO referral processing unit capacity, and referral processing time (at the MCO) were obtained from historical data and on-site observation study. Data cleaning and outliers analysis was performed in the same manner as described in Section 3.2.2. Then, data modeling was performed to fit the data to appropriate probability distributions so that they are incorporated into the DES model. General MCO tasks' processing times were obtained from an-onsite observation study. The minimum, mode, and maximum data points were readily available, and it was very difficult to obtain more data during this research course. Therefore, based on the available data, a triangular probability distribution was used for the referral processing tasks and activities in the MCO. Table 3 presents the model parameters and associated probability distributions and values.

### 3.3.2 Simulation Models Development

The simulation modeling part of the process improvement framework ensembles building two models: baseline and ML-Guided. The baseline model was developed to capture the behavior of the current system. Figure 5 shows the referral processing workflow which depicts the referral request journey and the interaction of the MCO referral processing unit with other entities such as inpatient hospitals, external providers, and insurance companies. It should be mentioned that this study focuses on the MCO and its affiliated hospitals. Therefore, the interaction between the MCO and the other entities was captured briefly to serve the feasibility of the current investigation. The DES model was implemented using Arena software.

Table 1. Summary of DES model parameters

| Parameter | Probability Distribution / Value |
| --- | --- |
| Referral arrival rate originating at $H_1$ [per day] | SNF: 20, HHS: 40, Other: 10 |
| Referral arrival rate originating at $H_2$ [per day] | SNF: 30, HHS: 40, Other: 10 |
| Referral arrival rate originating at $H_3$ [per day] | SNF: 30, HHS: 20, Other: 10 |
| Initial referral processing time [mins] | Triangular [3, 8, 12] |
| Waiting for more information from hospital [mins] | Triangular [5, 40, 120] |
| Sending request to vendor time [mins] | Triangular [10, 13, 15] |
| Waiting for vendor decision [days] | Triangular [0.125, 0.5, 6] |
| Sending authorization request internally [mins] | Triangular [3, 4, 5] |
| Authorization processing time [mins] | Triangular [15, 18, 20] |
| Waiting for insurance company decision [days] | Triangular [0.125, 0.85, 2] |
| Transportation arrangement [mins] | Triangular [20, 25, 30] |
| Referral request processing time at $H_1$ [days] | -0.5 + Lognormal [6.21, 4.72] |
| Referral request processing time at $H_2$ [days] | -0.5 + Lognormal [6.59, 5.55] |
| Referral request processing time at $H_3$ [days] | -0.5 + Lognormal [5.25, 5.29] |

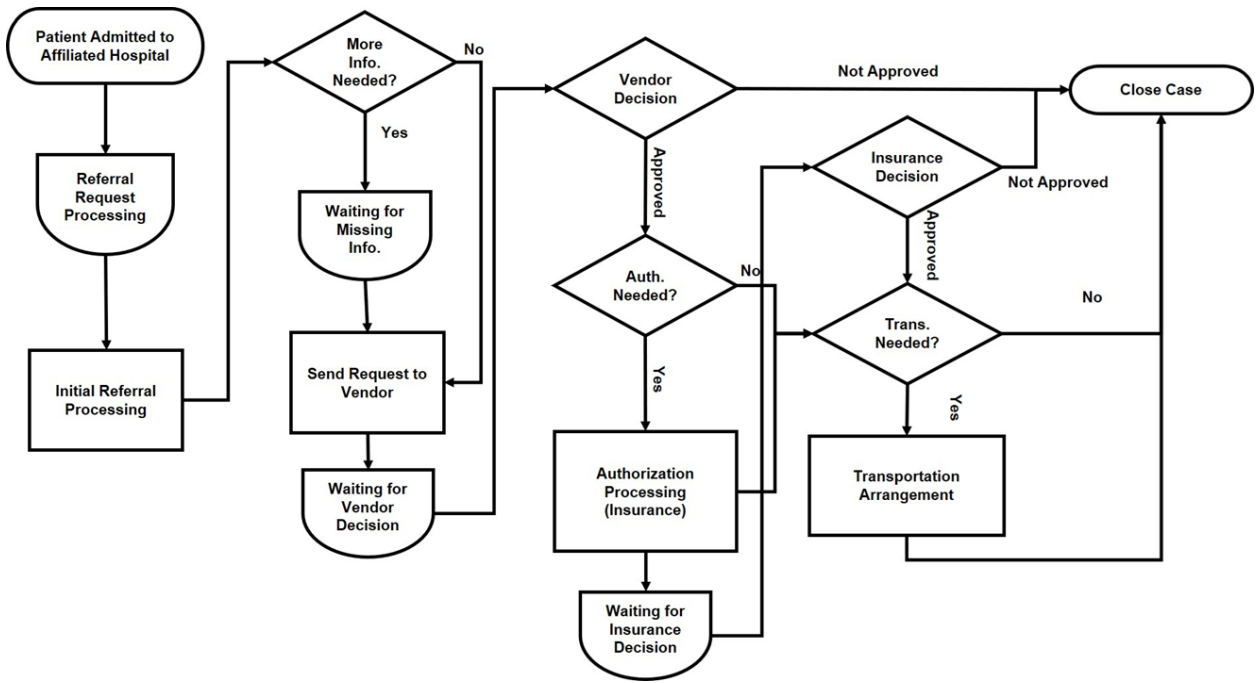

Figure 5. Health referral processing workflow

An important step in simulation modeling is to verify and validate the developed model against the real system to make sure that the degree of representation is satisfactory for the investigation goal. In addition to the programming logic and overall soundness verification, the developed model was checked against the actual process workflow to assess the degree of representation. The second round of validation was achieved by tracking a single referral request to study how logical the path taken by the request is. Finally, to obtain a quantitative measure of the degree of representation, historical data of the admission to referral time (ATRT) were collected. The data was obtained from the system repository. To balance the statistical comparison, one-month data was used for validation purposes. Similarly, the ATRT produced by the baseline DES model was captured. The probability density and cumulative density function of both variables are shown in Figure 6. It is noticed that they follow each ether quite closely, indicating a high degree of representation. The un-paired T-Test was used to compare the means of both variables: baseline ATRT and historical ATRT. The null hypothesis (i.e., the means are equal) failed to be rejected ($P = 0.52$).

The baseline model logic was altered to account for incorporating the machine learning models. First, the referral requests demand is predicted and fed into the DES model. Second, the queuing logic was changed from a first-come-first-served scheme to a priority scheme such that referrals are prioritized based on the patient's discharge date predicted by the machine learning model. Another inherent change was reflected at the hospital side as a reduction in referral request processing time based on an expert opinion. To illustrate, if the referral type and the discharge date is predicated upon admission (e.g., within the first day of admission), the inpatient hospital will need 2 days or less for requesting the referral from the MCO. A new triangular probability distribution was used (Triangular [0.5, 1, 2]) replacing the baseline values.

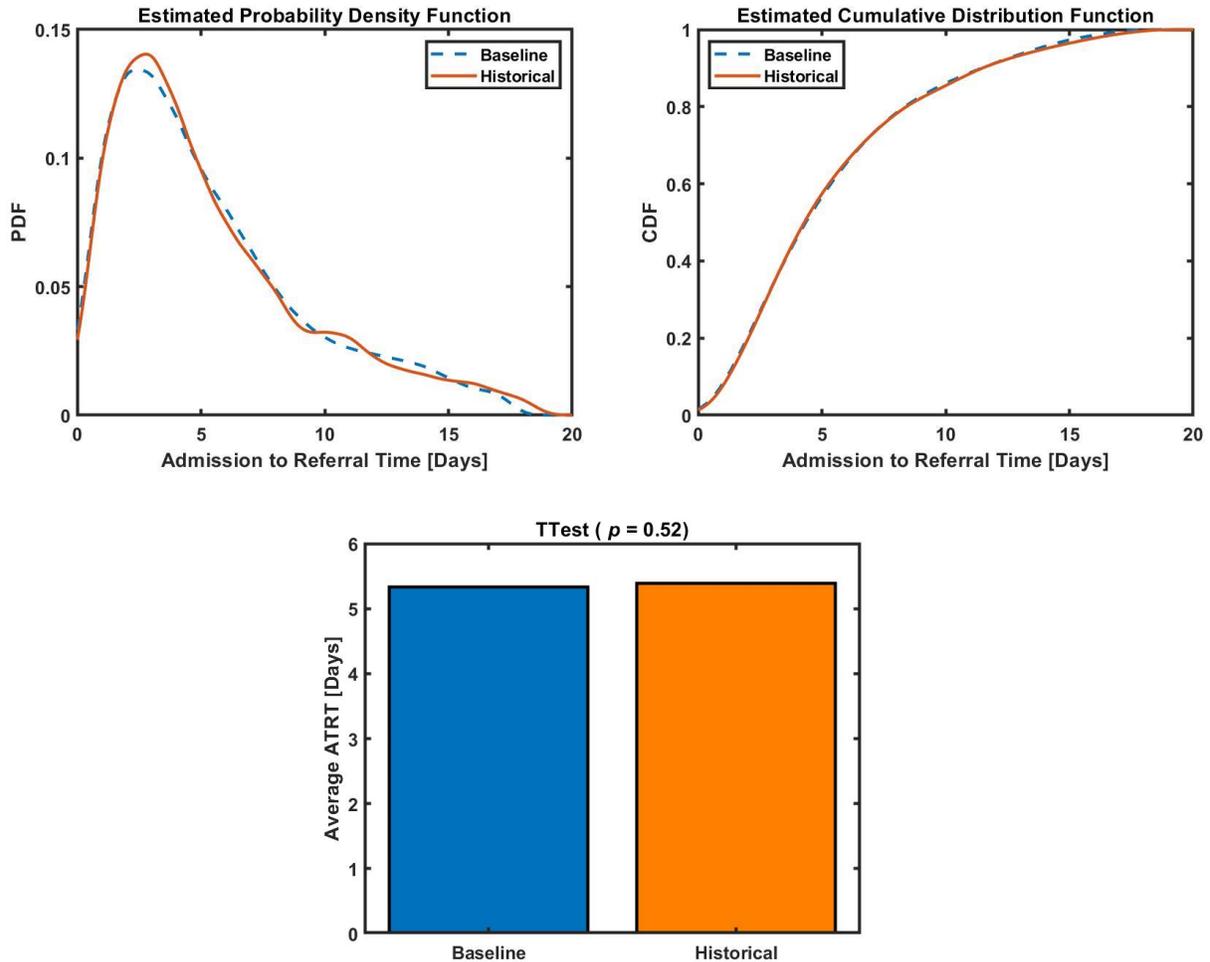

Figure 6. Baseline discrete event simulation validation

## 4. Results

This paper proposes an integrated process improvement framework based on machine-learning-guided (ML-Guided) simulation. The framework was applied to a case study from a care management setting. This section presents the results of the referral type prediction model, length of stay prediction (LOS) model, estimated demand, baseline discrete event simulation (DES) model, and the ML-Guided DES model.

To predict the referral type, a multi-class classification problem was solved using a random forest algorithm. The overall evaluation metrics of the referral type prediction model across all hospitals are presented in Table 2. The values in Table 2 ensemble the average of the 10 folds. The table shows the results of all affiliated hospitals. The multi-class classification model was applied to each hospital's dataset separately. This approach was adopted to reflect the actual system which is composed of several inpatient hospitals. The model was evaluated using four metrics: accuracy, AUROC, sensitivity, and specificity. Accuracy refers to the percentage of correct predictions concerning the total number of predictions performed. The AUROC is defined as the area under the receiver operating curve (ROC), which depicts the relationship between the true positive rate and the false positive rate (i.e., 1 – specificity). Sensitivity evaluates the model's true positive rate or the ability of the model to correctly classify the targeted referral type (e.g., SNF). On the other hand, specificity evaluates the model's ability to identify the negative. In this paper's context, when predicting SNF, the other referral types, which are considered as a negative class collectively, are HHS and others. It is very important to have a balanced model in terms of sensitivity and specificity.

In addition to the aggregated values, the actual distribution of the metrics plays a very important role in evaluating the performance of the prediction model. Figure 7 shows the distribution of the evaluation metrics across all hospitals. In Table 2, the aggregated sensitivity and specificity measures were presented. In other words, the sensitivity and specificity of individual referral types were averaged at each fold. However, it can be very insightful to assess the model's ability to identify a certain referral type, especially when that referral type is relatively more important. In Figure 7, the sensitivity and specificity distribution of individual referral types are depicted.

Table 2. Referral type prediction overall results across all hospitals

|       | Accuracy    | AUROC       | Sensitivity | Specificity |
|-------|-------------|-------------|-------------|-------------|
| $H_1$ | 90.15±0.83  | 97.94±0.30  | 90.17±1.93  | 95.08±0.83  |
| $H_2$ | 88.57±0.91  | 97.25±0.35  | 88.59±1.66  | 94.29±0.73  |
| $H_3$ | 89.35±1.84  | 97.59±0.58  | 89.34±2.81  | 94.68±1.43  |

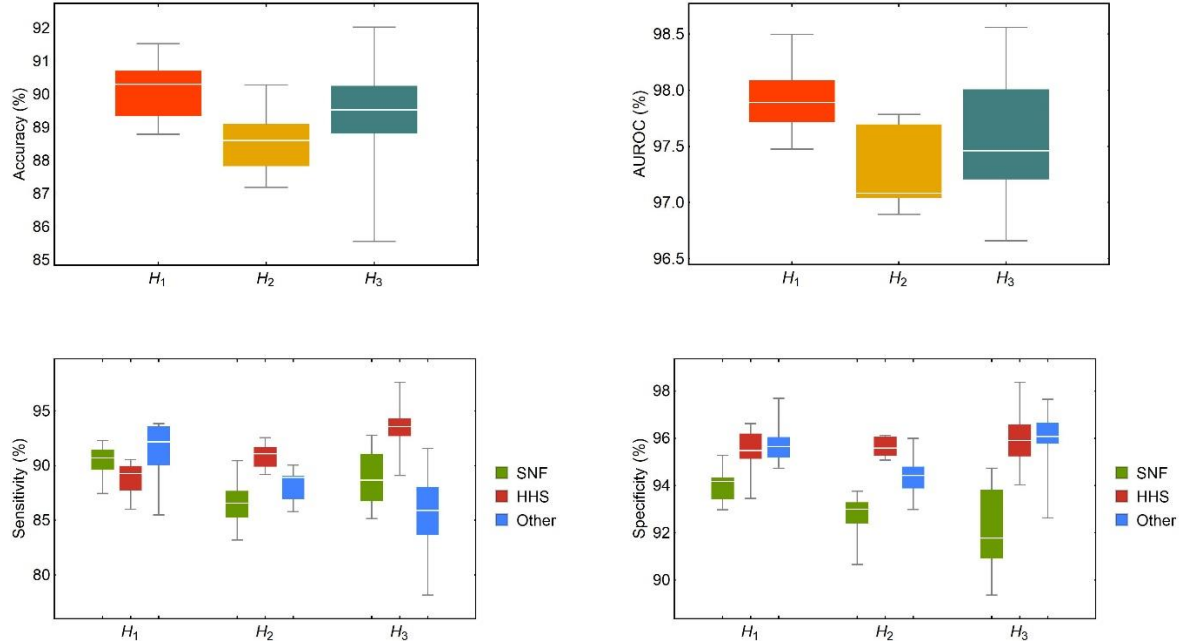

Figure 7. Referral type prediction evaluation metrics distribution

To predict the LOS, a regression problem was solved using random forest. The overall results of the LOS prediction models across all hospitals are presented in Table 3. The values in Table 3 represent the average of the 10 folds within the 10-fold cross-validation model evaluation approach. The regression model was applied to each affiliated hospital separately. This was adopted to reflect the actual system. The regression model was evaluated using three metrics: the mean absolute error (MAE), mean squared error (MSE), the coefficient of determination ($R^2$). The MAE refers to the average absolute difference between the actual values and predicted values. This metric does not penalize the errors. Therefore, other metrics should be calculated to obtain a complete picture of the model performance. The MSE is defined as the average of the squared difference between actual values and predicted values. $R^2$ is a score (less than 1) showing how the regression model performs compared to the baseline model. The baseline model in this paper was a naïve model predicting all instances as the average LOS in the testing sample. The actual distributions of these metrics are shown in Figure 8.

Table 3. LOS prediction overall results across all hospitals

|  | MAE | MSE | $R^2$ |
|---|---|---|---|
| $H_1$ | 1.90±0.12 | 7.96±0.97 | 0.67±0.04 |
| $H_2$ | 2.35±0.07 | 10.61±0.67 | 0.64±0.02 |
| $H_3$ | 2.52±0.36 | 11.75±2.41 | 0.50±0.09 |

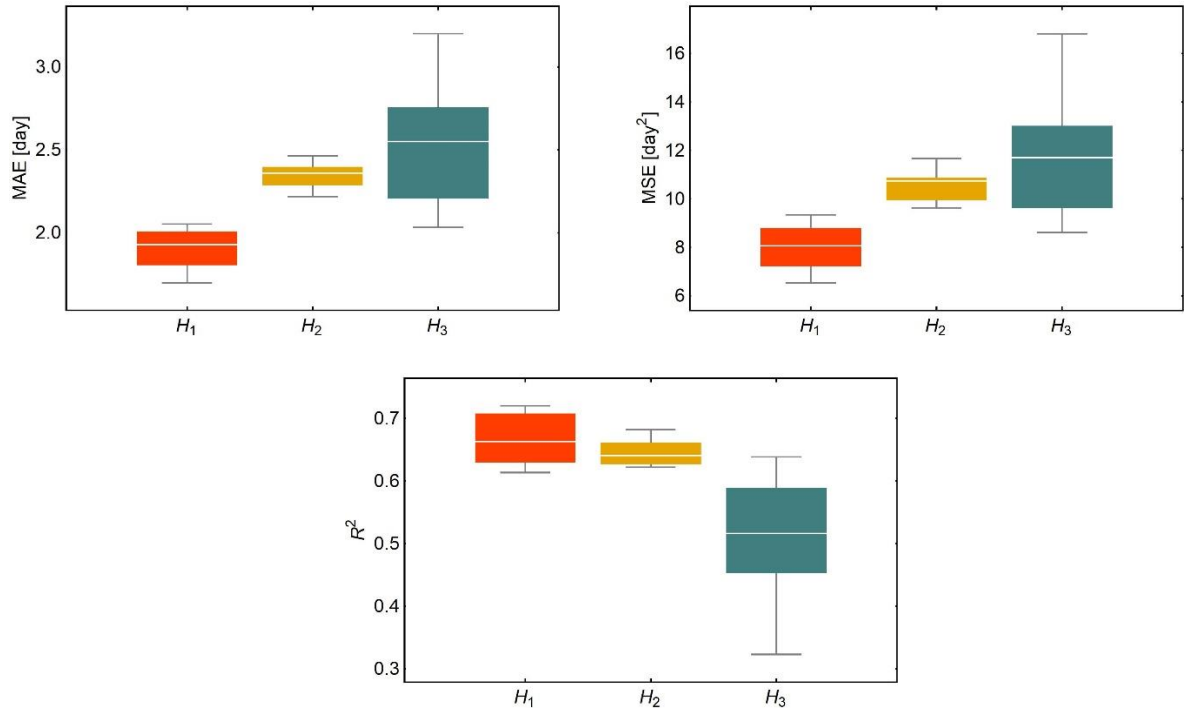

Figure 8. LOS prediction evaluation metrics distribution

The main objective of incorporating the prediction models in the process improvement framework is to estimate the patient discharge volume (i.e., the demand for the MCO referral processing unit) and to obtain a prioritization criterion. The prioritization criterion was obtained from the predicted LOS; the patient discharge date was calculated using the LOS value so that earlier discharge dates have higher priority. The estimated demand is displayed in Figure 9. The demand originated from each hospital is considered separately. This reflects the demanding nature of the real system. The estimated demand along with the prioritization criteria will be introduced to the ML-Guided simulation model.

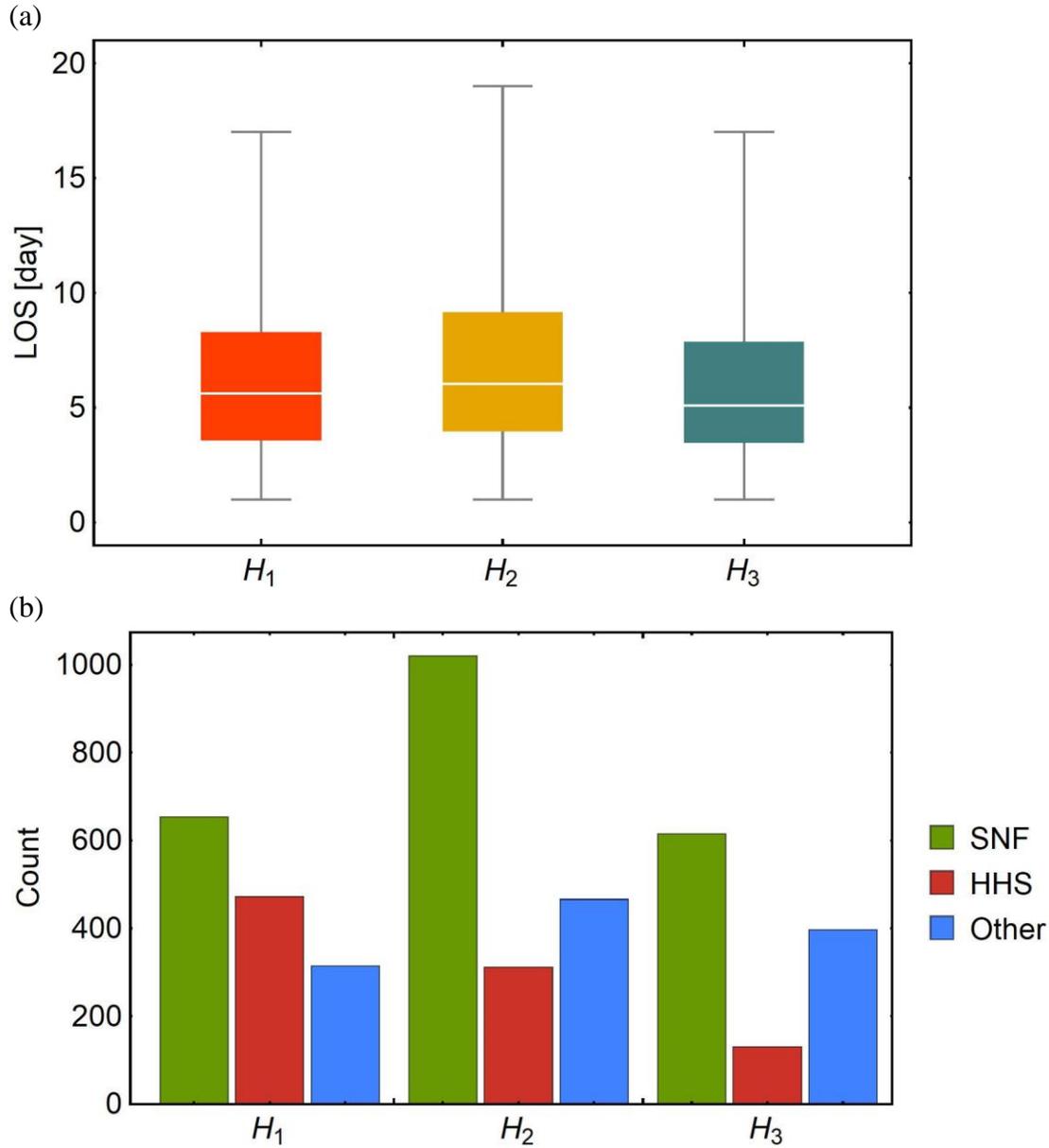

Figure 9. Estimated demand summary results (a) The predicted LOS distribution for a month (b) The referral type distribution for a month. The demand volume for a month for each hospital was determined according to historical data. The prediction models predicted the LOS and Referral Type of each patient.

The results of the baseline and ML-Guided simulation models are displayed in Figure 10 and Figure 11. Aligning with the goal of this paper, two performance metrics were captured: admission to referral time (ATRT) and referral creation delay time (RCDT). ATRT is defined as the number of days between patient admission to the affiliated inpatient hospital data and referral creation date. While RCDT is defined as the number of days between referral creation date and discharge date if and only if the referral creation data is larger than the discharge date. Otherwise, RCDT is zero.

These two metrics are used to evaluate the overall referral processing time and the process delay time. Figure 10 shows the distribution of ATRT of referrals originated at all affiliated hospitals for both the baseline and ML-Guided simulation models (n = 100). Figure 11 shows the distribution of RCDT of referrals originated at all affiliated hospitals for both the baseline and ML-Guided simulation models (n = 100).

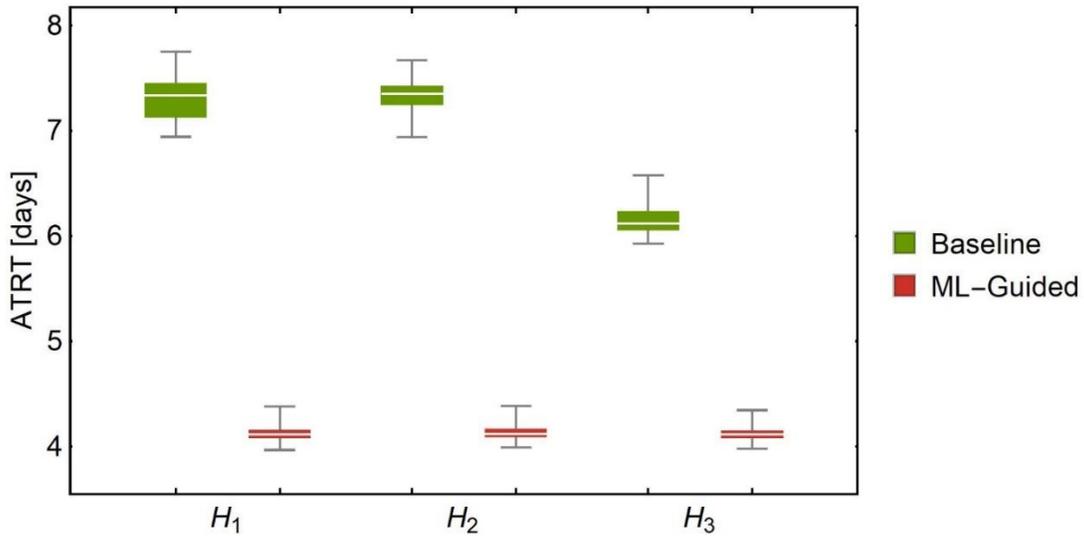

Figure 10. ATRT distribution across all hospitals.

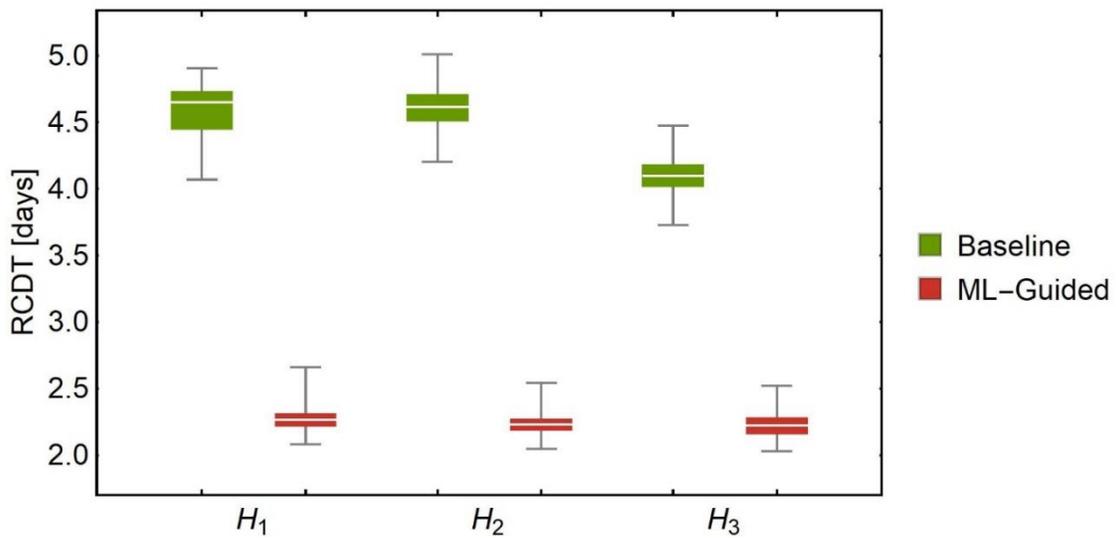

Figure 11. RCDT distribution across all hospitals.

## 5. Discussion

As mentioned in Section 2, most of the studies related to predicting post-discharge destination focused on a particular group of patients and there is a lack of studies tackling the problem of multi-class classification which reflects the nature of referral types in reality. Therefore, this paper solved a multi-class classification problem to predict the referral type (i.e., post-discharge destination) for comprehensive groups of patients. The random-forest-based referral type prediction model was performed across all the hospitals. The results varied a little from hospital to hospital. For instance, the $H_1$ prediction model outperforms other hospitals. In general, the prediction accuracy is larger than 88% for all models. This level of accuracy is satisfactory for discharge planning in a care management setting according to the expert opinion. However, depending on accuracy alone in classification problems can be very hazardous. Therefore, making sense of model sensitivity and specificity is very important. As shown in Table 2, the model performed well in terms of these two metrics. It was also shown that there is a good balance between the two metrics. The balance is manifested as high AUROC values ($\approx$ 97%). This indicates that the model can identify positive and negative classes to about the same degree of ability. A positive class in this context can be one referral type (e.g., SNF) and the negative class can represent all other referrals (e.g., HHS and Other). Looking at these two metrics from a granular point of view, the model performance with respect to predicting individual referral types varied across the referral types and the hospitals (see Figure 7).

As discussed in Section 2, the LOS prediction models in the literature focused on specific patient cohorts. The LOS model presented in this paper predicts the LOS for patients regardless of their associated cohort. This is believed to increase the validity of post-discharge planning. The random-forest-based LOS prediction model had good performance across all hospitals. It was shown that the regression model performed better on $H_1$ compared to other hospitals with respect to the evaluation metrics: MAE, MSE, and $R^2$. In general, the LOS model performance is satisfactory for discharge planning. One argument that can be made related to the $R^2$ is that prediction in healthcare business settings (like setting targeted by this paper) forecasting of continuous variables are made using naïve prediction. In other words, predictions are performed by calculating the average value. On the other hand, robust prediction models, similar to the one developed in this paper, can take the prediction accuracy to the next level. This can be manifested in the high $R^2$ values. To sum up, the LOS prediction model is way better compared to the baseline model.

After evaluating the prediction models, the discharge volume of each hospital was estimated from the predicted LOS. The distribution of the predicted LOS is depicted in Figure 9-a. The models can predict LOSs as high as 20. In addition to short-term planning, this can allow for long-term planning as well. However, if a patient is ought to stay longer, running the model regularly will overcome the shortcoming of this model. It can be noticed that the distribution of the LOS across all hospitals is very similar having an average value of about 5 days. Figure 9-b shows the distribution of predicted referral types across all hospitals. SNF has higher values in each hospital. This pattern is very similar to the original dataset. Moreover, $H_2$ has higher counts in total. This trend is also displayed in the original dataset. $H_2$ has a larger patient population compared to other hospitals.

To evaluate the current configuration of the referral processing integrated system, a baseline DES model was developed and evaluated. Then, the ML-Guided simulation model was built by changing the logic of the baseline model to incorporate the prediction capability that was used to estimate the discharge volume at the hospital side and to prioritize referral requests, based on the patient discharge date, at the MCO side. As discussed in Section 5, two evaluation metrics were used ATRT and RCDT. The distributions of these two metrics are depicted in Figures 10 and 11, respectively. It is shown that there is a huge improvement in both ATRT and RCDT when introducing the prediction capability to the system (i.e., ML-Guided simulation). This can be emphasized in Figures 12 and 13, which show a comparison between the baseline and ML-Guided models in terms of the average ATRT and RCDT. The ML-Guided simulation model reduced the overall referral processing time by about 42% for referral requests originating in $H_1$ and $H_2$. It also reduced the overall processing time by around 33% for referral requests originating in $H_3$. The average referral processing time reduction, by applying the proposed process improvement framework, for the whole system is nearly 39%. In terms of the delay time, the ML-Guided simulation model reduced the RCDT by 49% for referral requests originating in H1 and H2. It also reduced the RCDT by around 45% for referral requests originating in H3. The average RCDT reduction, by applying the proposed process improvement framework, for the whole system is nearly 48%.

To conclude, the proposed machine-learning-guided simulation approach could enhance the overall process of creating health referrals for patients admitted to affiliated hospitals. This was achieved by incorporating a prioritization criterion for referral requests based on the predicted patient discharge date. Besides, making the system smarter by predicting the referral type and patient discharge date, the hospital would become faster in sending the requests to the MCO referral processing unit, leading to lower overall processing times and fewer delays.

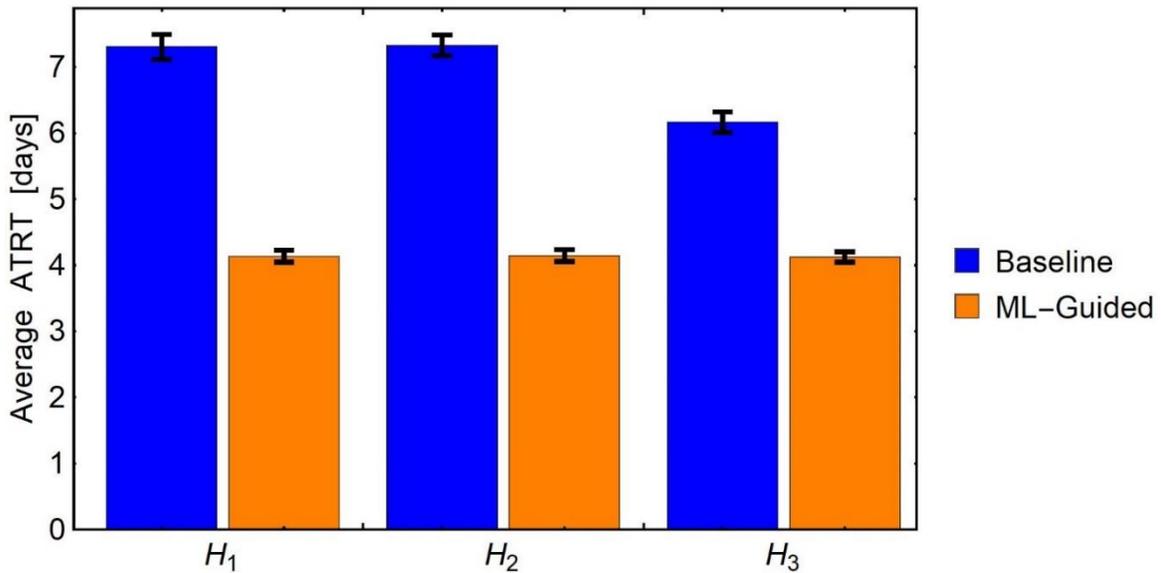

Figure 12. Average ATRT for baseline and ML-guided models across all hospitals

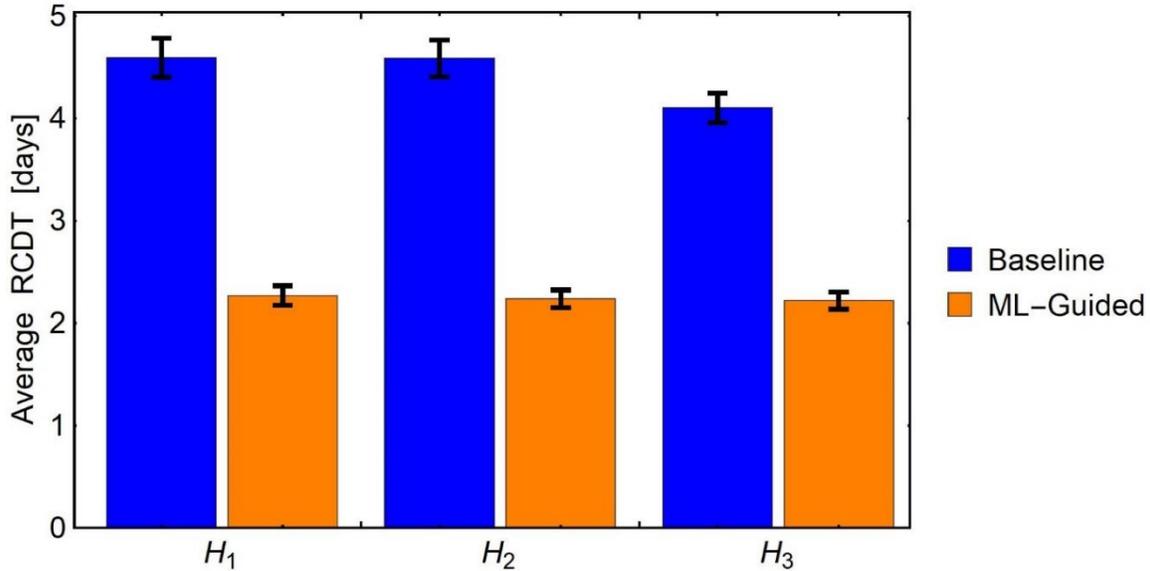

Figure 13. Average ATRT for baseline and ML-guided models across all hospitals

## 6. Conclusion

This paper presented a machine-learning-guided discrete event simulation framework to improve health referral processing in a care management setting. Two discrete event simulation models were developed: baseline and ML-Guided. Random-forest-based prediction models were used to predict the demand of the MCO referral processing unit for 30 days; this information was later introduced to the ML-guided DES model. The baseline DES model served as the foundation for validating against the actual system, identifying areas of improvement, and evaluating the performance of the current system. Incorporating the prediction functionality improved the overall performance substantially. Implementing the proposed framework reduced the processing and delay times significantly.

The process improvement framework developed in this paper can be used in the future to study and improve the system in many aspects. See below some examples:

i. Investigate the staffing level of the MCO referral processing unit to identify any bottleneck resulting from understaffing or reassess if overstaffing is the current status of the unit.
ii. The framework can also be used to test scenarios related to team specialization. For instance, to test the effect of introducing a specialized team for each referral type on the overall performance of the system.
iii. Evaluate the MCO referral processing unit's load if a new hospital is added to the current affiliations.

In general, this paper introduced an integrated process improvement framework that can be applied to studying and improving complex healthcare systems. Moreover, it is believed that this paper will emphasize the importance of applying integrated systems engineering methods in the healthcare domain, especially for coordinated healthcare systems (e.g., managed care systems).

**Conflict of interest**:
The authors declare that they have no conflict of interest.